\documentclass{article}
\usepackage[final]{nips_2018}
\usepackage{times}  
\usepackage{helvet}  
\usepackage{courier}  
\usepackage{url}  
\usepackage{graphicx}  
\usepackage{titlesec}
\usepackage{mathtools}

\titlespacing*{\paragraph}{0pt}{1.5ex plus 1ex minus .2ex}{1.3ex plus .2ex}
 
\usepackage{times}
\usepackage{xcolor}
\usepackage{soul}
\usepackage{nicefrac}
\usepackage{mathtools}

\usepackage{graphicx}
\usepackage{subfig}
\usepackage{caption}
\usepackage{amsmath} 
\usepackage{amssymb}
\usepackage{amsthm}
\usepackage{mathabx}
\usepackage{array}
\usepackage{bbm}
\usepackage[ruled,linesnumbered,vlined]{algorithm2e}
\usepackage{calrsfs}
\usepackage[capitalize]{cleveref}

\crefname{algocf}{Algorithm}{Algorithm}
\Crefname{algocf}{Algorithm}{Algorithm}

\DeclarePairedDelimiter{\norm}{\lVert}{\rVert}
\DeclarePairedDelimiter{\abs}{\vert}{\rvert}

\makeatletter

\newcommand*{\rom}[1]{\expandafter\@slowromancap\romannumeral #1@}
\makeatother
\newcommand{\E}{\mathbb{E}}
\renewcommand{\P}{\mathbb{P}}

\newcommand{\opt}{^\star}

\newcommand{\tr}{^{\mathsf{T}}}

\newcommand{\Real}{\mathbb{R}}
\renewcommand{\ss}{~:~}

\usepackage{natbib}
\newcommand{\RNum}[1]{\uppercase\expandafter{\romannumeral #1\relax}}

\newcommand{\states}{\mathcal{S}}
\newcommand{\actions}{\mathcal{A}}

\newcommand{\aset}{\mathcal{P}}
\newcommand{\aseth}{\aset^{H}}

\newcommand{\asetb}{\aset^{B}}

\newcommand{\RBU}{\widehat{T}}

\newcommand{\vset}{\mathcal{V}}

\newcommand{\statecount}{S}

\theoremstyle{plain}
\newtheorem{theorem}{Theorem}[section]

\theoremstyle{definition}
\newtheorem{definition}[theorem]{Definition}

\title{Tight Bayesian Ambiguity Sets for Robust MDPs}
\author{
Reazul Hasan Russel \hspace{16pt} Marek Petrik\\ 
Department of Computer Science\\
University of New Hampshire\\
Durham, NH 03824 USA\\
{\tt rrussel}, {\tt mpetrik} at {\tt cs.unh.edu}
}

\begin{document}
\maketitle


\begin{abstract}
Robustness is important for sequential decision making in a stochastic dynamic environment with uncertain probabilistic parameters. We address the problem of using robust MDPs~(RMDPs) to compute policies with provable worst-case guarantees in reinforcement learning. The quality and robustness of an RMDP solution is determined by its ambiguity set. Existing methods construct ambiguity sets that lead to impractically conservative solutions. In this paper, we propose RSVF, which achieves less conservative solutions with the same worst-case guarantees by 1) leveraging a Bayesian prior, 2) optimizing the size and location of the ambiguity set, and, most importantly, 3) relaxing the requirement that the set is a confidence interval.  Our theoretical analysis shows the safety of RSVF, and the empirical results demonstrate its practical promise. 
\end{abstract}

\section{Introduction}

Markov decision processes (MDPs) provide a versatile methodology for modeling dynamic decision problems under uncertainty~\citep{Bertsekas1996,Sutton1998,Puterman2005}.  MDPs  assume that transition probabilities are known precisely, but this is rarely the case in reinforcement learning. Errors in transition probabilities often results in policies that are brittle and fail in real-world deployments. A promising framework for robust reinforcement learning are robust MDPs (RMDPs) which assume that the transition probabilities and/or rewards are not known precisely. Instead, they can take on any value from a so-called \emph{ambiguity set} which represents a set of plausible values~\citep{Xu2006,Xu2009,Mannor2012,Petrik2012,Hanasusanto2013,Tamar2014a,Delgado2016,Petrik2016a}. The choice of an ambiguity set determines the trade-off between robustness and average performance of an RMDP.

The main contribution of this paper is RSVF, a new \emph{data-driven} Bayesian approach to constructing \emph{ambiguity} sets for RMDPs. The method computes policies with tighter safe estimates~(\cref{def:safety}) by introducing two new ideas. First, it is based on Bayesian posterior distributions rather than distribution-free bounds. Second, RSVF does not construct ambiguity sets as simple confidence intervals. Confidence intervals as ambiguity sets are a sufficient but not a necessary condition. RSVF uses the structure of the value function to optimize the \emph{location} and \emph{shape} of the ambiguity set to guarantee lower bounds directly without necessarily enforcing the requirement for the set to be a confidence interval.

\section{Problem Statement: Data-driven RMDPs} \label{sec:robust_mdps} 
We propose to use Robust Markov Decision Processes (RMDPs) with states $\states = \{1, \ldots, S \}$ and actions $\actions = \{1, \ldots, A \}$ to compute a policy with the maximal \emph{safe} estimate of return. 

\begin{definition}[Safe Estimate of Return] \label{def:safety}
	We say that an estimate of policy return $\tilde{\rho}: \Pi \rightarrow \Real$ is \emph{safe} with probability $\delta$ for a given dataset $\mathcal{D} \subseteq \{ (s,a,s') ~:~ s,s'\in\states, a\in \actions \}~,$ if it satisfies:
	$ \P_{P\opt} \bigl[ \tilde{\rho}(\pi) \le \rho(\pi,P\opt) ~\vline~ \mathcal{D} \bigr] \ge 1-\delta~,$
	for each stationary deterministic policy $\pi$. Here $P\opt: \states\times\actions\to\Delta^\states$ is the true, but unknown transition probabilities, and $\rho(\pi)$ is the return for a policy $\pi$.
\end{definition}

In standard \emph{batch} RL setting, $\mathcal{D}$ can be used to estimate the transition probabilities, but is assumed to be not known precisely for the RMDP and is constrained to be in the \emph{ambiguity set} $\aset_{s,a}$, defined for each state and action (s,a-rectangular). The most common method for defining ambiguity sets is to use norm-bounded distance from a \emph{nominal} probability distribution $\bar{p}$:
$
\aset_{s,a} = \{p \in \Delta^\statecount \ss \norm{p - \bar{p}_{s,a} }_1 \le \psi_{s,a} \}
$ 
for a given $\psi_{s,a}\ge 0$ and a nominal point $\bar{p}_{s,a}$. We assume that the rewards $r:\states\times\actions\to\Real$ are known. The \emph{objective} is to maximize the $\gamma$-discounted infinite horizon return.

\begin{figure}
	\centering
	\begin{minipage}[c]{.45\columnwidth}
		\centering
		\includegraphics[width=0.8\linewidth]{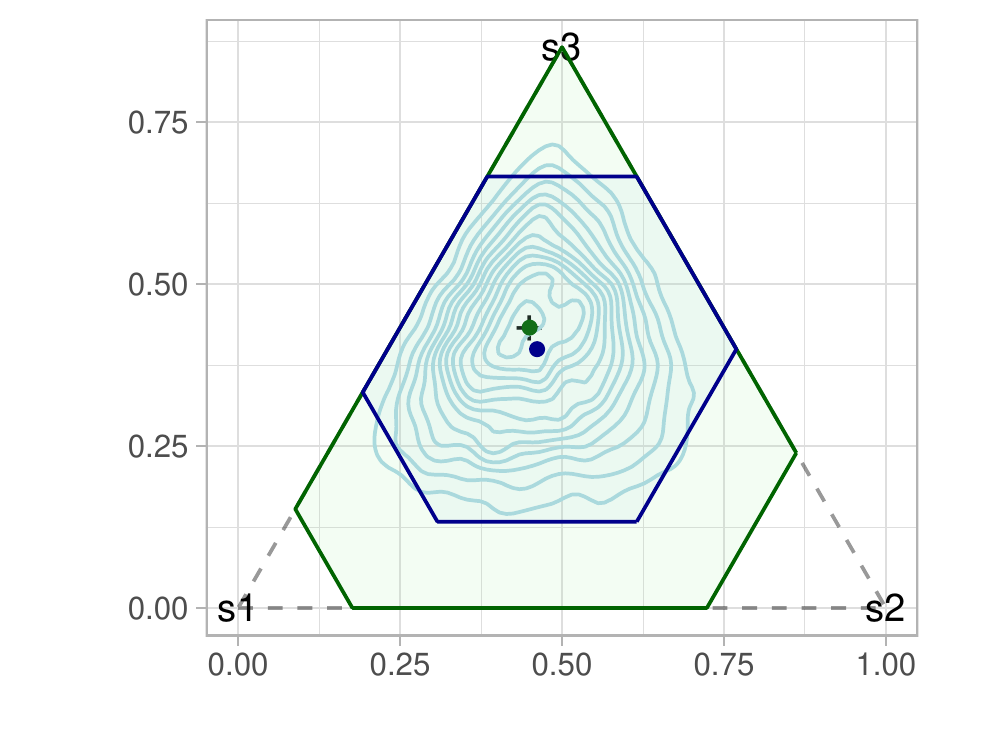}	\\
	\end{minipage}%
	\begin{minipage}[c]{.45\columnwidth}
		\centering
		\includegraphics[width=0.8\linewidth]{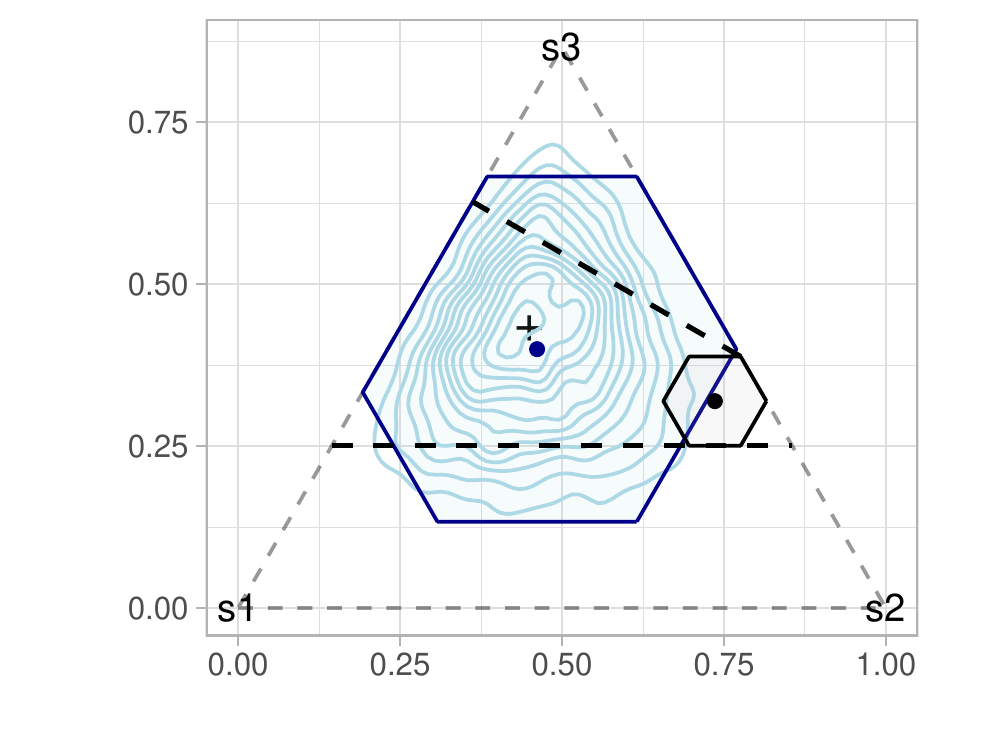} \\
	\end{minipage}%
	\caption{Comparison of 90\% $L_1$ confidence intervals, Left: Hoeffding (green) and Bayesian (blue), Right: RSVF (green) and BCI (blue).}
\end{figure}

RMDPs satisfy similar properties as regular MDPs~\citep{Iyengar2005,Tamar2014a}. The robust Bellman operator $\RBU_\aset$ is defined for a state $s$ as:
$ 
(\RBU v)(s) := \max_{a\in\actions}\min_{p \in\aset_{s,a}}  (r_{s,a} + \gamma \cdot p\tr v)  
$.
The robust return is defined as~\citep{Iyengar2005}:
$ \hat{\rho}(\pi) = \min_{P\in\aset} \rho(\pi, P) = p_0\tr \hat{v}^\pi~, $
where $p_0 \in \Delta^S$ is the initial distribution. In general, we use hat $(\hat{.})$ to denote quantities in RMDP.

\section{Ambiguity Sets as Confidence Intervals} \label{sec:confidence_interval}
In this section, we describe the standard approach to constructing ambiguity sets as distribution-free confidence intervals and propose its extension to the Bayesian setting. 

\textbf{Distribution-free Confidence Interval} The use of distribution-free error bounds on the $L_1$ norm is common in reinforcement learning~\citep{Petrik2016a,Taleghan2015,Strehl2004}. The confidence interval is constructed around the mean transition probability by combining the Hoeffding inequality with the union bound~\citep{Weissman2003xx,Petrik2016a}. The Hoeffding ambiguity set is defined as:
$ 
\aseth_{s,a} = \left\{ \norm{p_{s,a}\opt - \bar{p}_{s,a} }_1 \le \sqrt{\frac{2}{n_{s,a}} \log \frac{S A 2^{S}}{\delta} } \right\}
$
where $\bar{p}_{s,a}$ is the mean transition probability computed from $\mathcal{D}$ and $n_{s,a}$ is the number of transitions observed originating from state $s$ and an action $a$. An important limitation of $\aseth$ is that the size of the ambiguity set grows linearly with the number of states $S$.

\textbf{Bayesian Confidence Interval (BCI)}
Here we assume that data $\mathcal{D}$ is available and a hierarchical Bayesian model can be used to infer a probability distribution over $P\opt$ analytically or using MCMC methods like Stan~\citep{Gelman2004}. To construct the ambiguity set $\asetb$, we optimize for the \emph{smallest} ambiguity set around the mean transition probability with the assumption that a smaller ambiguity set will lead to a tighter lower bound estimate. Formally, the optimization problem to compute $\psi_{s,a}$ for each state $s$ and action $a$ is:
$
\min_{\psi\in\Real_+} \left\{\psi \ss \P\left[ \norm{p\opt_{s,a} - \bar{p}_{s,a}}_1 > \psi ~|~ \mathcal{D} \right] < \frac{\delta}{SA} \right\}~,
$
where nominal point is $\bar{p}_{s,a} = \E_{P\opt}[p\opt_{s,a} ~|~ \mathcal{D}]$.

\begin{figure}
	\centering
	\begin{minipage}[c]{.45\columnwidth}
		\centering
		\includegraphics[width=\linewidth]{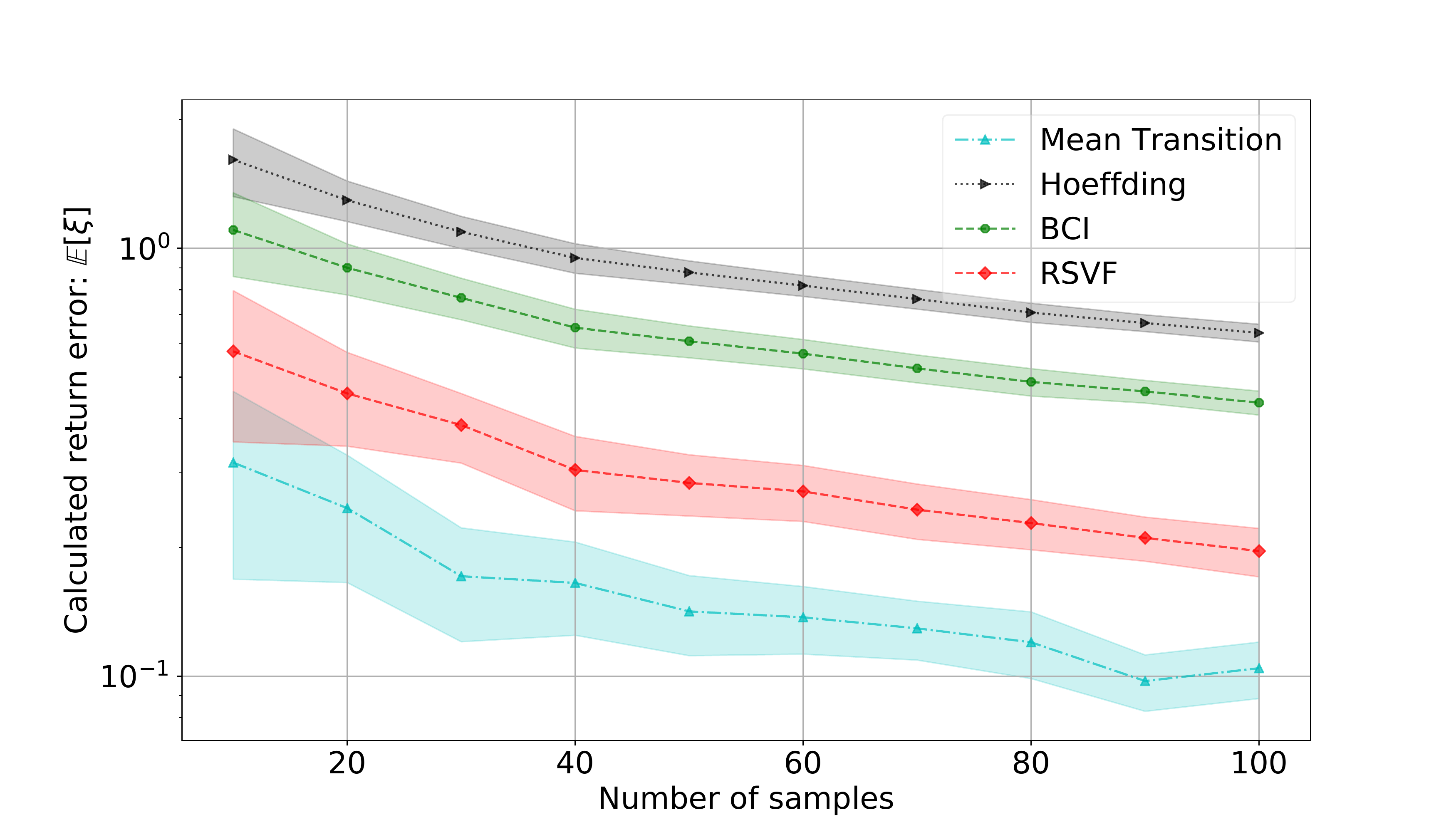}
	\end{minipage}%
	\begin{minipage}[c]{.45\columnwidth}
		\centering
		\includegraphics[width=\linewidth]{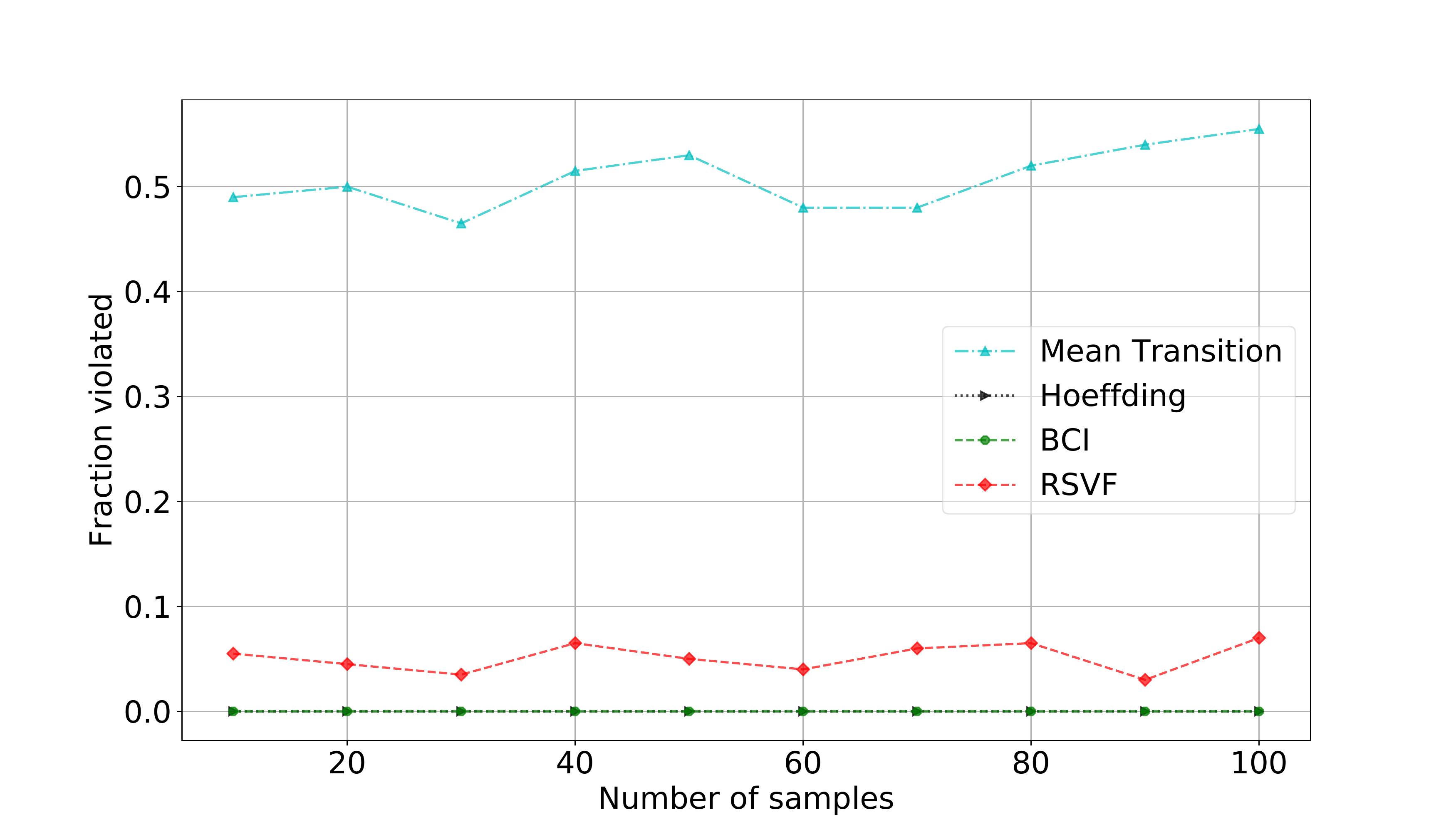}
	\end{minipage}%
	\caption{Single state with Dirichlet prior, return error with 95\% Confidence \& violations.}
	\label{fig:dirichlet_result}
\end{figure}

\section{RSVF: Robustification With Sensible Value Functions} \label{sec:multiple}

RSVF uses samples from a posterior distribution, similar to a Bayesian confidence interval, but it relaxes the safety requirement as it is sufficient to guarantee for each state $s$ and action $a$ that:
\begin{equation} \label{eq:condition_safe}
\min_{v\in\mathcal{V}} \P_{P\opt} \left[ \min_{p \in \aset_{s,a}} (p - p_{s,a}\opt)\tr v \le 0  ~\middle|~ \mathcal{D} \right] \ge 1-\frac{\delta}{SA}~,
\end{equation}
with $\mathcal{V} = \{ \hat{v}\opt_{\aset} \}$. To construct the set $\aset$ here, the set $\mathcal{V}$ is not fixed but depends on the robust solution, which in turn depends on $\aset$. RSVF starts with a guess of a small set for $\mathcal{V}$ and then grows it, each time with the current value function, until it contains $\hat{v}\opt_{\aset}$ which is always recomputed after constructing the ambiguity set $\aset$.

\begin{algorithm}
	\KwIn{Desired confidence level $\delta$ and posterior distribution $\P_{P\opt}[\cdot ~|~\mathcal{D}]$ }
	\KwOut{Policy with a maximized safe return estimate }
	Initialize current policy $\pi_0 \gets \arg\max_{\pi} \rho(\pi,\E_{P\opt}[P\opt~|~\mathcal{D}])$\;
	Initialize current value $v_0 \gets v^{\pi_0}_{\E_{P\opt}[P\opt~|~\mathcal{D}]}$\;
	Initialize value robustness set $\mathcal{V}_0 \gets \{v_0 \}$ \;	
	\label{line:make_p_1} Construct $\aset_0$ optimal for $\mathcal{V}_0$\;
	Initialize counter $k\gets 0$\;
	\While{\cref{eq:condition_safe} is violated with $\mathcal{V}=\{v_k\}$}{
		Include $v_k$ that violates \cref{eq:condition_safe}: $\vset_{k+1} \gets \vset_k \cup \{ v_k \}$ \;
		\label{line:make_p_2} Construct $\aset_{k+1}$ optimized for $\vset_{k+1}$\;
		Compute robust value function $v_{k+1}$ and policy $\pi_{k+1}$ for $\aset_{k+1}$\;
		$k \gets k + 1$ \;
	}
	\Return $(\pi_k, p_0\tr v_k)$ \;
	\caption{RSVF: Robustification with Sensible Value Functions}    \label{alg:IAVF}
\end{algorithm} 

In lines 4 and 8 of \cref{alg:IAVF}, $\aset_i$ is computed for each state-action $s,a \in \states\times\actions$. Center $\bar{p}$ and set size $\psi_{s,a}$ are computed from \cref{eq:center_point} using set $\mathcal{V}$ \& optimal $g_v$ computed by solving \cref{eq:optimal_hyperplane}.
When the set $\mathcal{V}$ is a singleton, it is easy to compute a form of an optimal ambiguity set. 
\begin{equation} \label{eq:optimal_hyperplane}
g = \max \left\{ k ~:~ \P_{P\opt} [k \le v\tr p\opt_{s,a}] \ge 1 - \delta/(SA) \right\}
\end{equation}

When $\mathcal{V}$ is a singleton, it is sufficient for the ambiguity set to be a subset of the hyperplane $\{ p \in \Delta^S ~:~ v\tr p = g\opt \}$ for the estimate to be safe. When $\mathcal{V}$ is not a singleton, we only consider the setting when it is discrete, finite, and relatively small. We propose to construct a set defined in terms of an $L_1$ ball with the minimum radius such that it is safe for every $v\in\mathcal{V}$. Assuming that $\mathcal{V} = \{v_1, v_2, \ldots, v_k \}$, we solve the following linear program:
\begin{equation} \label{eq:center_point}
\begin{gathered}
\psi_{s,a} = \min_{p\in\Delta^S} \Bigl\{ \max_{i=1,\ldots,k} \norm{q_i - p}_1 ~:~ 
\hspace{0.1cm}  v_i\tr  q_i = g_i\opt, q_i \in \Delta^S, i \in 1,\ldots,k  \Bigr\}
\end{gathered}
\end{equation}

In other words, we construct the set to minimize its radius while still intersecting the hyperplane for each $v$ in $\mathcal{V}$. \cref{alg:IAVF}, as described, is not guaranteed to converge in finite time as written. It can be readily shown the value functions in the individual iterations are non-increasing. It is easy to just stop once the value function becomes smaller (and that is more conservative) than BCI.

\begin{figure}
	\centering
	\begin{minipage}[c]{.45\columnwidth}
		\centering
		\includegraphics[width=\linewidth]{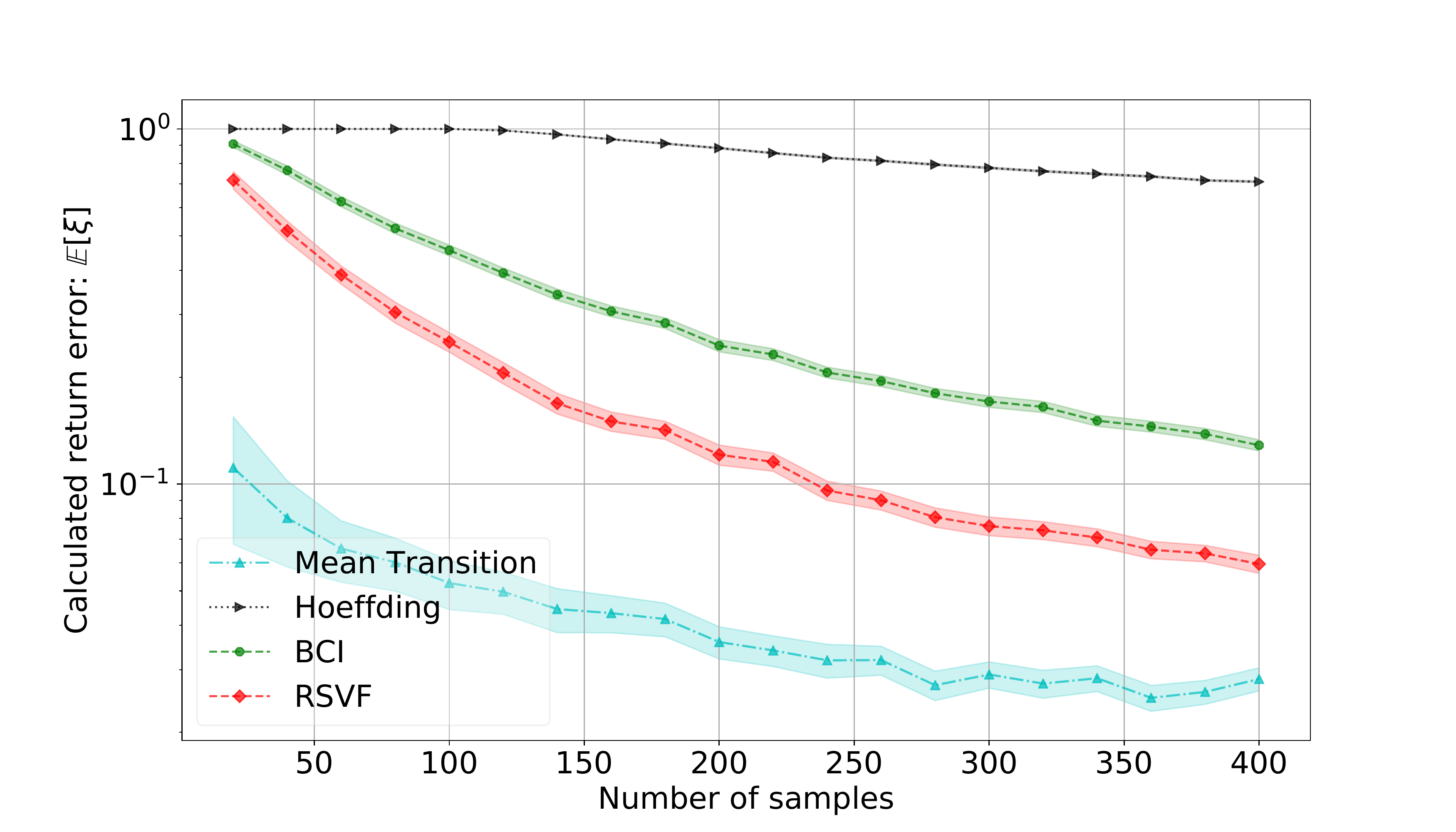}\\
	\end{minipage}%
	\begin{minipage}[c]{.45\columnwidth}
		\centering
		\includegraphics[width=\linewidth]{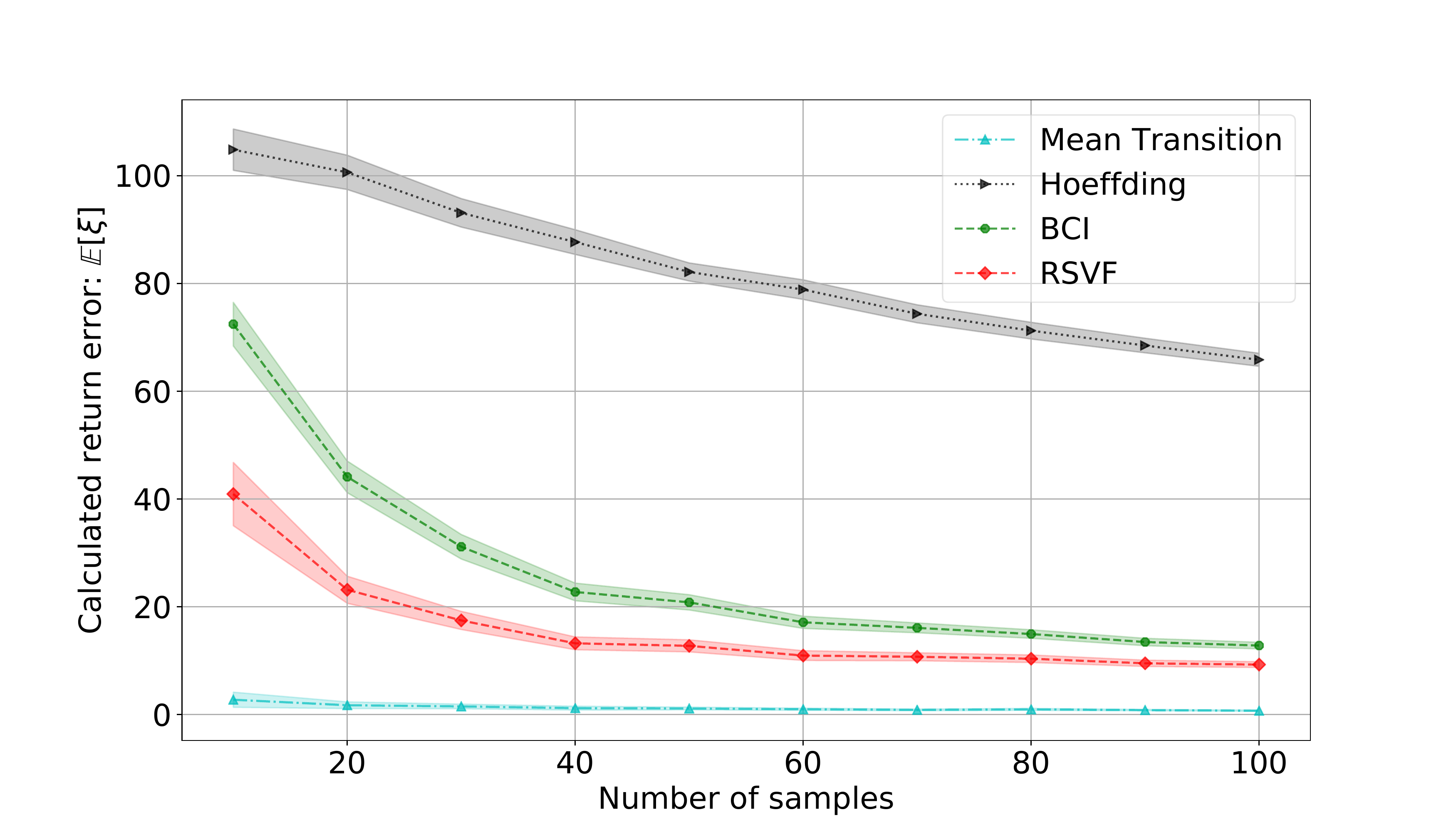}
	\end{minipage}%
	\caption{Return error with a Gaussian prior with 95\% confidence, Left: Single state, Right: Full MDP, X-axis is the number of samples per state-action.}
	\label{fig:return_single_multiple}
\end{figure}

\section{Empirical Evaluation} \label{sec:experiments}

In this section, we evaluate the safe estimates computed by BCI and RSVF empirically. We assume a true model of each problem and generate a number of simulated data sets for the known distribution. We compute the largest safe estimate for the optimal return and compare it with the optimal return for the true model. We compare our results with ``Hoeffding Inequality`` based distance $\aseth$ and ``Mean Transition'' which simply solves the expected model $\bar{p}_{s,a}$ and provides no safety guarantees. The value $\xi$ represents the predicted regret, which is the absolute difference between the \emph{true} optimal value and the robust estimate: $\xi = \abs{\rho(\pi\opt_{P\opt}, P\opt) - \hat\rho(\hat\pi\opt)}$, a smaller regret is better. All of our experiments use a 95\% confidence for safety unless otherwise specified.

\textbf{Single-state Bellman Update} We initially consider simple problems where transition from a single non-terminal state following a single action leads to multiple terminal states. The value function for the terminal states are fixed and assumed to be provided. We evaluate different priors over the transition probabilities: i) uninformative Dirichlet prior and ii) informative Gaussian prior. Note that RSVF is optimal in this simplistic setting, as \cref{fig:dirichlet_result} (left) and \cref{fig:return_single_multiple} (left) shows. As expected, the mean estimate provides the tightest bound, but \cref{fig:dirichlet_result} (right) illustrates that it does not provide any meaningful safety guarantees.

\textbf{Full MDP with Informative Prior} Next, we evaluate RSVF on a full MDP problem. Standard RL benchmarks, like cart-pole or arcade games, lack meaningful Bayesian priors. We instead use a simple exponential population model, based on the management of an invasive species~\citep{Taleghan2015}.
The population $N_t$ of the invasive species at time $t$ evolves according to the exponential dynamics $N_{t+1} = \min{(\lambda_t N_t, K)}$.  Here, $\lambda$ is the growth rate and $K$ is the carrying capacity of the environment. A land manager needs to decide, at each time $t$, whether to take a control action which influences the growth rate $\lambda$. If $z_t$ is the indicator of whether the control action was taken, the growth rate $\lambda_t$ is defined as: 
$\lambda_t = \bar{\lambda} - z_t N_t\beta_1 - z_t\max{(0, N_t-\bar{N})^2}\beta_2 + \mathcal{N}(0,\sigma_y^2)$, 
where $\beta_1$ and $\beta_2$ are the coefficients of control effectiveness. We also assume that we only observe $y_t$, a noisy estimate of population $N_t$:  $y_t \sim N_t + \mathcal{N}(0,\sigma_y^2)$. In the MDP model, the population observation defines the state. There are two actions: to apply or not to apply the control measure. Transition probabilities are given by the population evolution function. The reward for the MDP captures the costs of high invasive population and the application of the treatment. 

\cref{fig:return_single_multiple} (right) depicts the average predicted regret over the different datasets. The distribution-free methods are very conservative, BCI improves on this behavior somewhat, but RSVF provides bounds that are even tighter than BCI by almost a factor of 2. The rate of violations is 0 for all robust methods. This indicates that RSVF is overly conservative in this case too since its rate of violations is also close to 0. This is due its reliance on the union bound across multiple states, the approximate construction of the individual ambiguity sets, and the inherent rectangularity assumption.

\section{Conclusion}
We propose, in this paper, a new Bayesian approach to the construction of ambiguity sets in robust reinforcement learning. This approach has several important advantages over standard distribution-free methods used in the past. Our experimental results and theoretical analysis indicate that the Bayesian ambiguity sets can lead to much tighter safe return estimates. 

\bibliography{marek,reazullib}

\newpage
\appendix
\onecolumn

\end{document}